\documentclass{article}

\usepackage{spconf,amsmath,graphicx}
\usepackage{url}
\usepackage{todonotes}

\title{Detecting Emotion Carriers by Combining Acoustic and Lexical Representations}

\name{Author Name$^1$, Co-author Name$^2$}
\name{Sebastian P. Bayerl$^1$ Aniruddha Tammewar $^2$ Korbinian Riedhammer $^1$ Giuseppe Riccardi $^2$}

\address{$^{1}$ Technische Hochschule Nürnberg Georg Simon Ohm, Germany\\$^{2}$ Signals and Interactive Systems Lab, University of Trento}

\begin{document}

\maketitle
\begin{abstract}
Personal narratives (PN) -- spoken or written -- are recollections of facts, people, events, and thoughts from one’s own experience.
Emotion recognition and sentiment analysis tasks are usually defined at the utterance or document level.
However, in this work, we focus on Emotion Carriers (EC) defined as the segments (speech or text) that best explain the emotional state of the narrator ({\tt"loss of father"}, {\tt"made me choose"}).
Once extracted, such EC can provide a richer representation of the user state to improve natural language understanding and dialogue modeling.
In previous work, it has been shown that EC can be identified using lexical features.
However, spoken narratives should provide a richer description of the context and the users' emotional state.
In this paper, we leverage word-based acoustic and textual embeddings as well as early and late fusion techniques for the detection of ECs in spoken narratives.
For the acoustic word-level representations, we use Residual Neural Networks (ResNet) pretrained on separate speech emotion corpora and fine-tuned to detect EC. 
Experiments with different fusion and system combination strategies show that late fusion leads to significant improvements for this task.
\end{abstract}

\section{Introduction}\label{sec:introduction}
\begin{figure*}[t]
    \centering
    \includegraphics[width=\textwidth]{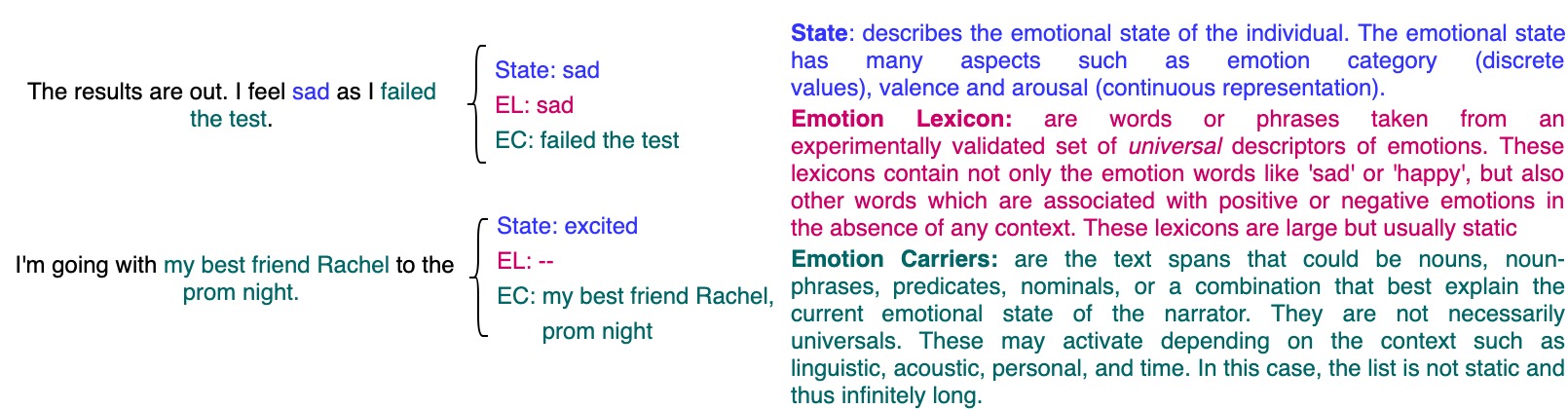}
    \caption{This figure explains and differentiates the concepts of \textit{Emotion State, Emotion Lexicon, and Emotion Carriers}. In the first example the emotional state 'sad' is directly described by the emotion lexicon using the word 'sad', whereas the second example does not use emotion lexicon to describe the emotional state 'excited', it is rather implicit and can be explained using emotion carriers.}
    \label{fig:emotion_tasks}
\end{figure*}
People express and communicate emotions consciously as well as subconsciously.
This is done by modifying the manner of speaking, the content of a conversation or written text, facial expressions, gestures, or even the way of walking. The combination of these signals, especially speech and text, has successfully been used to determine the emotional state of a person making a statement or telling a narrative~\cite{montacie2018vocalic,poria_review_affective_fusion_2017}. Most of the emotion research in affective computing aims at categorical or continuous recognition of emotions.
However, these tasks are not necessarily able to provide an explanation for the emotional state.
Tammewar \emph{et al.} defined the linguistic {\em carrier} of emotions and evaluated an annotation protocol in the context of PNs \cite{tammewar2020annotation}.
Detecting the fragments from the narratives that best explain the emotional state of the narrator would provide deeper emotion analysis that has potential benefits in the context of mental well-being applications, to analyze a user's Personal Narratives (PN).
This application of EC is described in \cite{tammewar2020emotion}.
A conversational agent, as part of a mental well-being application, uses previously extracted ECs, from the dialogue with the user, and generates a tailored response.
A deeper analysis of the the perceived emotion may help to ask better questions and get more information from the user.
This additional information in turn may help to better understand a user's emotional state through a conversation and makes showing empathy towards the user easier.

A previous analysis on how different fragments of PNs help explain the current mental state of the narrator in terms of valence  can be found in \cite{tammewar_modeling_2019}. 
Not only fragments containing emotion words but also events ({\emph ``high school exam''}), people ({\emph ``grandpa''}), and actions ({\emph ``made me choose''}) were proven to be useful to predict the valence.
Following up on this analysis, \emph{Emotion Carriers (EC)} for PNs can be defined as the concepts that best explain and carry the emotional state of the narrator \cite{tammewar2020annotation}.
ECs thus include not only explicitly emotionally charged words, such as {\emph ``happy''}, but also mentions of people, places, objects, predicates, and events that carry an emotional weight within the context of a narrative.
They performed the annotation of German PNs from the Ulm State-of-Mind in Speech (USoMs) corpus \cite{rathner2018state} with the emotion carriers \cite{tammewar2020annotation}. In Figure~\ref{fig:emotion_tasks}, we explain and differentiate the three different emotion concepts: \textit{Emotion State, Emotion Lexicon}, and \textit{Emotion carriers}.

Further work on the automatic detection of emotion carriers from transcriptions of spoken PNs, can be found in \cite{tammewar2020emotion}.
However, relying only on lexical features leaves out the possibility of the same lexical content conveying different things based on acoustic context. 

Ivanov \emph{et al.} showed that there is a relationship between meaning-bearing parts of utterances and their acoustic properties \cite{ivanov_acu_sem_2010}.
Following up on that research, we have found evidence supporting distinct prosodic profiles for emotion {\em vs} non-emotion carriers:
\figurename~\ref{fig:ec_spectrum} compares the spectrograms of two occurrences of the phrase {\emph ``vor die Wahl gestellt''; ``made me choose''}; while (a) was annotated as an emotion carrier, (b) was not.
The strong rise in fundamental frequency (f0), as well as the strong fluctuations at the beginning of \figurename~ \ref{fig:ec_spectrum}a, indicates emotional speech \cite{paeschke_f0_emo_1999}.
In contrast, the same phrase that was not marked as an EC has a very flat f0 contour (\emph{cf.} \figurename~\ref{fig:ec_spectrum}b).
While the figure provides only anecdotal and motivational evidence, in this paper we provide ample evidence of the complementarity of acoustic and lexical information.

Our contributions are:
\begin{itemize}
    \item evidence for the acoustic discriminability of EC
    \item word-level acoustic embeddings using a modified ResNet architecture and transfer learning from EmoDB
    \item analysis of early and late fusion of textual and acoustic embeddings
    \item rule-based late fusion based on posterior probabilities leveraging the strength of the lexical system
    
\begin{figure*}[htb]
\begin{minipage}{0.5\linewidth}\label{fig:res_a}
  \centering
  \centerline{\includegraphics[width=8.0cm]{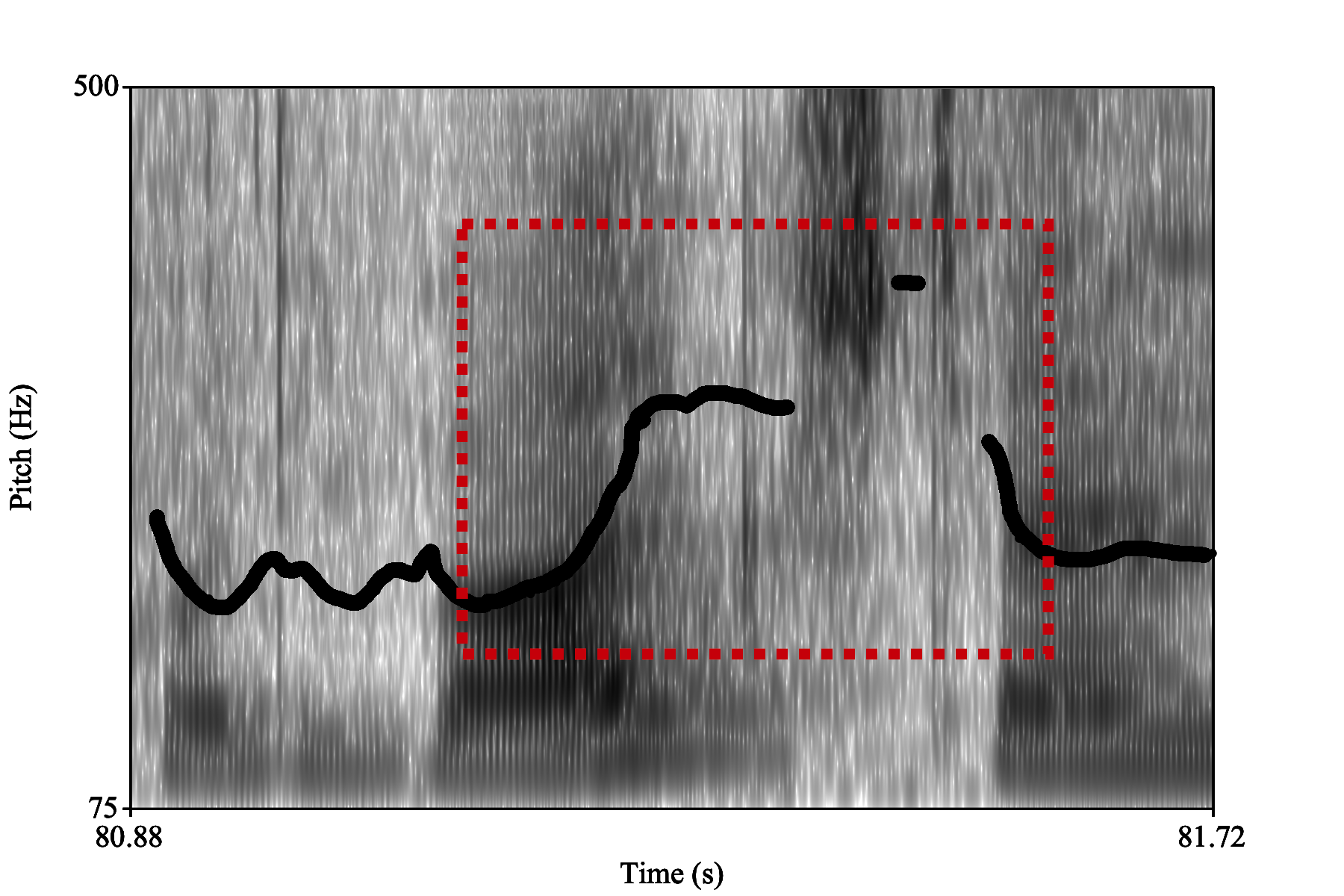}}
  \centerline{(a) emotion carrier}\medskip
\end{minipage}
\begin{minipage}{0.5\linewidth}\label{fig:res_b}
  \centering
  \centerline{\includegraphics[width=8.0cm]{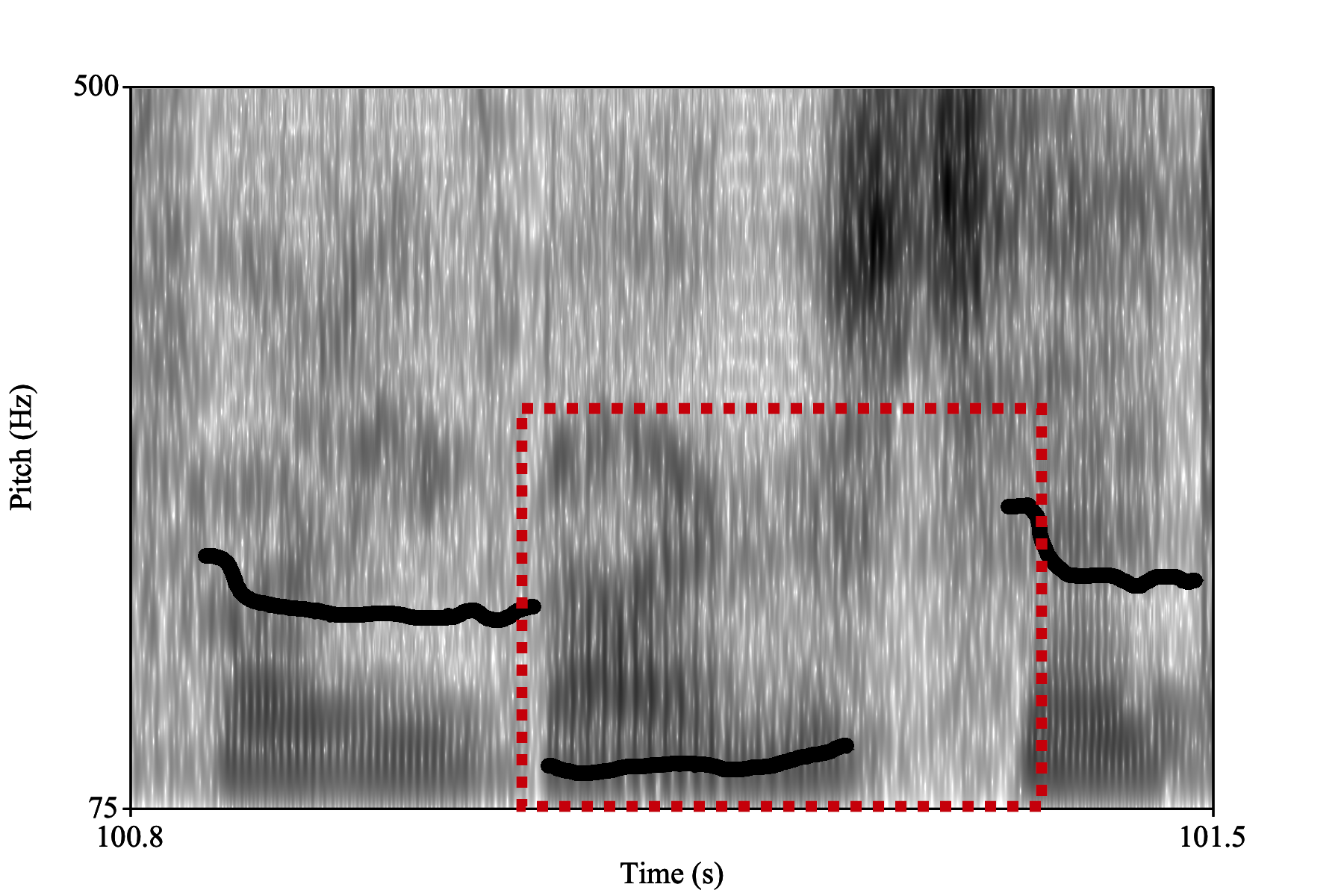}}
  \centerline{(b) non emotion carrier}\medskip
\end{minipage}
\caption{
Spectrograms with \textit{f0}-contour of the phrase: ``\textbf{vor die Wahl gestellt},''(Translation: ``made me choose'').
(a) was marked as an EC, showing signs of emotional speech and the voice cracking in the center part whereas (b) was taken from the same recording session, but was not marked as EC.
While this is an anecdotal example, statistical analysis revealed significant differences in f0, energy, and shimmer on nouns marked as EC when comparing them with all other nouns in the dataset. 
}
\label{fig:ec_spectrum}
\end{figure*}

\end{itemize}
\section{Related Work}
\label{sec:relatedwork}
A related concept to EC are \emph{affective events} (AE)
~\cite{ding2016acquiring}. 
Affective events are a predefined set of events that have a stereotipycally (w/o specific context) positive or negative impact on the people who experience the event.
In the set of AEs, an event is a tuple of subject, verb and object and the polarity is provided to the entire tuple, while ECs could be any entity or event, with no strict restriction on the syntactic categories, that carry the emotional state of the narrator.
AEs are closely linked to the satisfaction or violation of human needs ~\cite{ding_affective_event_2018}. 
While this can be true for EC, it is not a necessity, as EC can be identified in personal narratives, where only the recollection of events leads to the current emotional state.
Identifying what moves or creates an emotional reaction in a personal narrative can for example help a personalized mental health application ask questions about what is important w.r.t. the emotional state of the narrator. 

To our knowledge, there is no prior work on merging acoustic and lexical features for  detecting AE.

Relevant techniques for utilizing multiple modalities can be found in the related field of Speech Emotion Recognition (SER), where the use of multiple modalities is common \cite{pepino_fusion_2020,georgiou_deep_2019,hazarika_conversational_2018,meng_speech_2019}.
Most research on Emotion Recognition (ER) is focused on classifying the emotion of a writer or narrator into discrete values such as fear, anger, or joy.
This is usually done on an utterance, dialogue, or narrative level \cite{schuller_ser_2018}.
  
The SER task is concerned with the direct classification of a persons' emotional state whereas EC detection is looking for events, entities, and people that explain the subject's current emotional state.
ECs are the linguistic {\em vehicles} of emotions and provide insights into the entities, people, and events that explain the narrators' emotional state.
Related work focusing on acoustic cues in meaning structures can be found in \cite{ivanov_acu_sem_2010} where the authors found a relationship between acoustics and meaning. 
Nastase \emph{et al.} studied the relation between words that express emotions and the way they sound.
The study found statistical evidence that phonetic features are useful in determining if words express the same emotion \cite{nastase_happy_2007}.
Batliner \emph{et al.} used word-level emotional labels and tried classifying emotional state of children on word and turn level, using acoustic features as well as a combination of lexical and acoustic features \cite{batliner_combining_2006}.
Huang \emph{et al.} combine word-level acoustic embeddings with Word2Vec for utterance-level emotion recognition \cite{huang2018ser}


Current fusion approaches for multi-modal emotion recognition essentially follow a similar approach. 
The first step is finding appropriate intermediate representations, usually produced by using some kind of encoder neural network, and then, depending on the problem, train a classifier applying different fusion strategies \cite{pepino_fusion_2020,georgiou_deep_2019,hazarika_conversational_2018,meng_speech_2019}. 
For example, Pepino \emph{et al.} use CNNs to extract sentence-level embeddings from pre-trained word embeddings and utterance-level embeddings using a CNN on handcrafted acoustic features comparing different fusion approaches \cite{pepino_fusion_2020}.
\section{Data}\label{sec:data}

All experiments presented to detect EC and analyses in this paper were conducted using the dataset and annotations described in \cite{tammewar2020annotation}.
The dataset is based on the USoMs corpus described in the Interspeech 2018 ComParE paralingiustics challenge \cite{schuller_interspeech_2018}. 
The USoMs dataset consists of spoken PNs collected from 100 participants; audio data was converted to 16\,kHz mono, and a noise profile was removed where necessary.
Manual transcriptions were obtained from a professional transcription service (verbatim approach \cite{dresing_2011}), with a vocabulary of 6438 words.
The data of 66 participants (239 PNs) was annotated with the ECs by four annotators, selecting emotion-carrying text spans using the transcript only \cite{tammewar2020annotation}. 
ECs are annotated with the inside–outside (IO) scheme which is common in natural language processing.  
Words that belong to a text span marked as EC are annotated as \textit{I} others as \textit{O}.
The resulting dataset is heavily imbalanced with only 6.6\,\% of tokens marked as EC.



Accurate time alignment of text to audio was produced by forced alignment (FA) using a speaker-adaptive HMM-GMM (Hidden Markov Model, Gaussian Mixture Model) automatic speech recognition system (ASR) based on \cite{milde_tuda_2018}.
Missing entries in the pronunciation lexicon were generated using a grapheme-to-phoneme tool \cite{bisani_g2p_joint_seq_mod_2008} where necessary.

Before running any classification and fusion experiments for the detection of EC, we performed an in-depth feature analysis. 
We focused on prosodic word-level features of nouns marked as EC, as only they were labeled in a significant number of cases being an EC ($N=15600$, of which 2019 were EC nouns).
While Fig.~\ref{fig:ec_spectrum} provides anecdotal evidence only, we could identify significant differences between the nouns marked as EC and those that were not.
F0, energy, and HNR (Harmonic to Noise Ratio) were extracted using Praat with the Parselmouth library and Jitter and Shimmer were calculated based on the extracted f0 using the method described in \cite{farrus_using_jitter_2009}.
Our analysis revealed significant differences on a word-level for mean f0 (and its derivatives), mean energy (and its derivatives) as well as shimmer using an independent two-sample t-test ($p=0.05$).



\section{Method}
We follow the approach of finding representations for EC from different feature spaces. 
As EC are a word-, and phrase-level concept, we try to find appropriate representations from the linguistic and acoustic input feature space.
The representations are then used in uni-modal experiments as well as multi-modal fusion experiments for EC recognition.

\begin{figure*}[t]
    \centering
    \includegraphics[width=\linewidth]{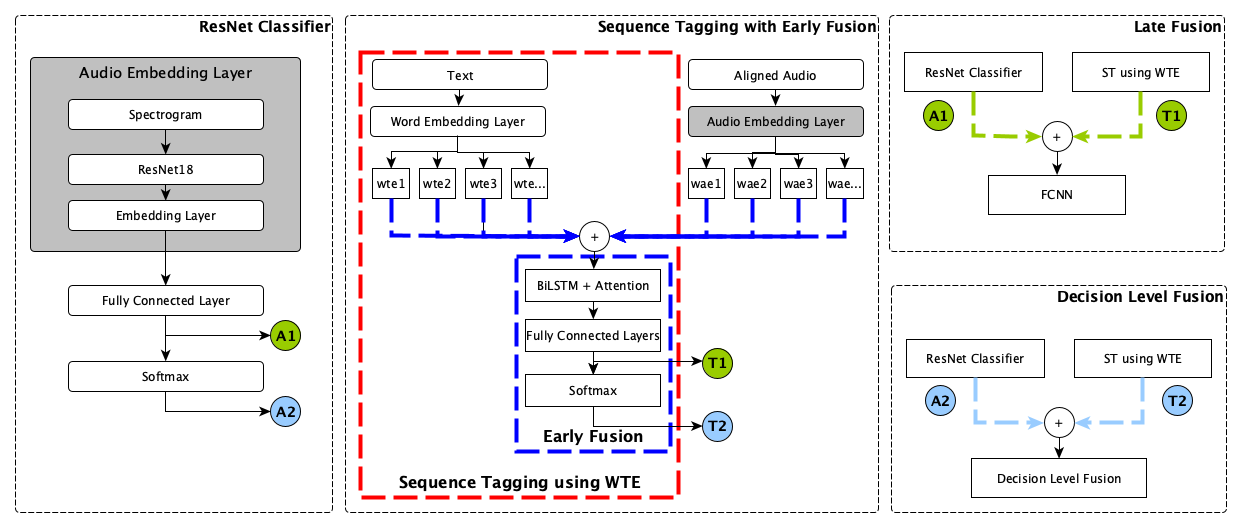}
    \caption{
    Overview of the neural network architecture used in the experiments.
    The left part shows the ResNet classifier containing the audio embedding layer which is used to extract word-based acoustic embeddings (WAE) for each word.
    The central part contains the sequence tagging (ST) architecture that can operate using either word-based textual embeddings (WTE), WAE, or a combination of WTE and WAE in an early fusion (EF) approach.
    The right part of the figure is depicting the late fusion (LF) and decision level fusion (DLF) systems. Inputs for the LF are taken after the fully connected layer in the ResNet and the ST, using only WTE as inputs, as logits (A1, T1) and for DLF after the Softmax layer (A2, T2) returning normalized probabilities.
    }
    \label{fig:architecture}
\end{figure*}

\subsection{Word-based textual embeddings (WTE)}
For word-based textual embeddings, we use 100-dimensional pre-trained GloVe word embeddings trained on the German Twitter corpus \cite{pennington_glove_2014}. 
A total of 656 (10.2\%) words were not present in the pre-trained embeddings.
The word-embeddings where fine-tuned on the actual task inside the cross-validation loops.

\subsection{Word-based acoustic embeddings (WAE)}

The previously performed feature analysis revealed differences between EC and non EC with respect to handcrafted acoustic features. 
This motivated us to use embeddings based on convolutional neural networks (CNN) and handcrafted acoustic features as described in \cite{pepino_fusion_2020}.
Our early stage experiments failed to produce good results on an utterance level with the German EmoDB dataset \cite{burkhardt_emodb_2005} on the speech emotion recognition (SER) task.

The failure to produce good results with CNNs and handcrafted acoustic features led us to explore other network architectures and input features.
We decided to use a ResNet architecture which was successfully applied to a number of speech applications such as speaker recognition, and SER and has been shown to produce good embeddings \cite{Chung_2018_voxceleb, tang_2018_resnet_ser}.
ResNets are fully convolutional neural networks (FCN) and can handle inputs of different sizes (or lengths, respectively) due to a global pooling layer at the end of the convolutional part of the neural network.
The network we use is very similar to the one described in \cite{he_resnet_2016} and consists of 18 convolutional blocks (ResNet18).
Its architecture was adapted by removing the initial max-pooling layer to keep more features prior to the residual blocks as the expected inputs are already relatively small.
The dimensionality of the embedding layer was reduced from 1000 to 512 and the final classification layer was altered to match the number of classes (2).

As acoustic input features for the ResNet, we extract 40-dimensional Mel-frequency cepstrum coefficients (MFCC) with a window length of 0.025\,s, a frameshift of 0.01\,s, along with 1st and 2nd order moments, stacking them to a tensor with three dimensions (frequency x time x moments) and apply z-score normalization. 
Those acoustic input features are then used to train the acoustic only classifier and to extract neural acoustic word embeddings from the trained acoustic encoder.
The word-based acoustic contexts are extracted using the aforementioned FAs.

The network was pre-trained on short utterances from the German EmoDB corpus to differentiate between neutral and emotional speech \cite{burkhardt_emodb_2005}. 
This is done to learn filters that are already primed to extract features from speech that are important to classify emotional speech. 
Both pre-training and training were done using stochastic gradient descent with a cross entropy loss function.
To overcome the class imbalance problem, an oversampling strategy was applied as it had proven to be the best performing technique in our experiments.

To obtain word-based acoustic embeddings (WAE), we froze to resulting model to act as an acoustic word encoder. 
We feed the word-based acoustic context to the model and extract WAE from the embedding layer of the model. 

\subsection{Sequence Tagging}
\label{sec:sequence_tagging}
We model the task of detecting emotion carriers as a binary sequence labeling problem using both modalities, with targets encoded as \textit{I} if the token is part of an EC and \textit{O} otherwise.
For this task, we adopt a bidirectional Long Short-Term Memory (LSTM) neural network with an attention-based sequence tagging (\textbf{ST}) architecture previously used for labeling candidates for emphasis in written text \cite{shirani2019learning}.\footnote{Implementation: \url{https://tinyurl.com/seqtagging}}
An overview of the architecture is located in the central part of Fig.~\ref{fig:architecture}. 

\subsection{Fusion}
There are two main challenges in combining multiple modalities: How to combine features of different dimensionality and valuation, and at which stage to combine the streams.
In general, three different kinds of approaches can be differentiated: early, late, and decision-level fusion. 
\subsubsection{Early and late fusion}
In early fusion (EF), features for each modality are extracted separately, i.e.~each modality represents a view of the same concept.
The resulting feature vectors are then combined, e.g.~by concatenation or stacking, and then treated as a single input channel.
In late fusion (LF), each modality has its own model and is often trained independently.
The outputs of those classifiers are used as input to another classifier that combines them for an overall best prediction.
EF as used in this paper can be found at the center of Figure~\ref{fig:architecture} and the LF approach can be found at the top right. 

\subsubsection{Decision Level Fusion}\label{sc:DLF}


In our experiments the sequence tagger using WTE is trained as a regression problem with the Kullback–Leibler (KL) divergence, predicting the probability of a token being an emotion carrier. 
For this, the best decision threshold was experimentally found to be $p_{db}=0.15$ for lexical features only. 
This motivated us to explore a rather heuristic late fusion approach: a \textit{rule-based cascaded classifier based on posterior probabilities.}
Applying a similar technique to the normalized probabilities in the output of the ResNet classifier, we can find a decision threshold and then merge the decisions, defining decision states around these thresholds.
This way it is possible to leverage long-range lexical information as well as local acoustic information. 

We define the lexical-based ST model to be the primary model and the ResNet classifier to be the disambiguator, leveraging local acoustic information.
In our merging approach, we define an $\epsilon$ parameter that indicates how certain a classifier is with the decision if a token is an EC.
The decision boundary (DB) is defined by setting a probability value $p_{db}$.

We only consider the probability for the EC to determine certainty.
If the normalized probability of a token being an EC $p_{ec}$ is within the epsilon interval $(p_{db} \pm \epsilon)$ the classifier is considered to be uncertain regarding a positive decision of a token being an EC.
$p_{ec} > p_{db} + \epsilon$ is considered to be certain.
Those certainty indicators are computed for both models separately.
Merging is then done by checking certainty indicators:
If the lexical model is certain, the token is considered to be an EC.
If the lexical model is uncertain and the acoustic model is certain, the token is also considered as an EC.
In all other cases, the token is not considered as an EC.
We call this heuristic \textit{decision level fusion} (DLF).

\section{Experiments}\label{sec:experiments}

Results for single modality, fusion experiments, and baselines, are reported in Tab.~\ref{tbl:results}.
As this is an information retrieval task, we report metrics for class \textit{I} in this unbalanced task.
The equal priors baseline constitutes random guessing with no knowledge about the actual class distribution with $p_{I}=p_{O}=0.5$, resembling a fair coin toss whereas the class priors baselines resembles a heavily biased coin with $p_{I}=0.066$ and $p_{O}=0.934$. 
\subsection{Training Details}
All experiments were performed using five-fold cross-validation with consistent folds across all experiments.
The folds were split by speaker to ensure no speaker in the test set was present in the training set and hyper-parameters were tuned on separate development folds, as is common when working with acoustic data and small datasets.
Tab.~\ref{tbl:results} contains results for single modality classification using a ResNet classifier as well as ST using either WTE or WAE as inputs.
We report one result for an EF experiment concatenating WTE and WAE to a single word vector as well as a logit-based LF experiment combining the ResNet classifier and the ST using WTE as inputs only.
Lastly, we show our overall best results, obtained with DLF and oracle results.
Oracle results are obtained by a fictitious fusion of classifiers, which is considered to be right, if at least one of the contributing classifiers (ResNet18 and ST WTE), predicted the correct label.
Details of the proposed neural network architectures can be found in Fig.~\ref{fig:architecture}.
\subsection{Results}

\begin{figure}[htb]
  \centering
  \includegraphics[width=0.8\linewidth]{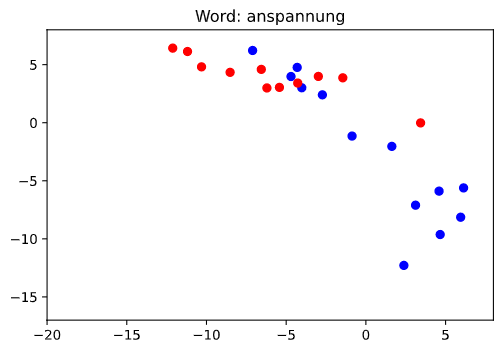}
\caption{t-SNE plot for the word-based acoustic embeddings of the German word "Anspannung" (English: tension). Red dots represent tokens marked as EC while blue are non EC.}
\label{fig:ec_tsne}
\end{figure}

Direct word-level EC detection using only MFCC features (ResNet18) improved results compared to both random baseline classifiers using class priors for both actual class priors in the dataset as well as equal class priors.
It can therefore be assumed that useful representations can be extracted from the embedding layer of the ResNet classifier.
The analysis of the word-based acoustic embeddings produced by the ResNet system also looked promising.
Fig.~\ref{fig:ec_tsne} contains a t-distributed Stochastic Neighbor Embedding (t-SNE) plot of embeddings marked as EC vs. embeddings not marked as EC. 
The plot shows that there is potential to differentiate EC from non-EC tokens in this low-dimensional projection.

While we achieved good results with the ST using WTE only, results for the WAE failed to perform better than the ResNet classifier that solely relied on local acoustic information.
It barely improved results compared to the random classifiers' expected baseline precision.
We, therefore, decided to not use the ST with WAE in late fusion experiments and rather use the ResNet classifier in LF.


The EF experiment combining WAE and WTE performed worse than WTE alone as described in this paper and only slightly improved previous WTE only results \cite{tammewar2020emotion}. 
The LF experiment using logit outputs from the ResNet classifier and the ST using WTE improved the ST using WTE only in terms of recall, but lowered the precision which lead to overall worse results w.r.t. F1-I.
Experiments with Logistic Regression to model the probability of a word being an EC using the logit outputs of the ST using WTE and the ResNet classifier did not improve over the LF experiment with the FCNN. 

Unfortunately, the experiments using the standard EF and LF approaches couldn't improve over the already strong textual system (ST WTE).
However as shown in Fig.~\ref{fig:ec_spectrum} and the feature analysis, there definitely is evidence that acoustic information can help with the detection of EC.
Our experiments with the word-level ResNet classifier could not completely convince but still beat all statistical baselines as a stand-alone system.
Lastly oracle results presented in Tab.~\ref{tbl:results} show that the combination of the ResNet classifier and the ST using WTE has a lot of room for improvements still.

This led us to explore the rather heuristic decision level fusion (DLF) approach described in ~\ref{sc:DLF} and yielded the best overall results. 
The decision boundary was tuned for the ResNet classifier only since the ST using WTE was already trained using KL divergence with a tuned decision boundary at $p_{db}=0.15$.
The DB for the ResNet classifier was determined using 5-fold cross-validation.
Results are reported in Tab.~\ref{tbl:results} (DLF).
The decision boundary for certainty of the ResNet classifier was found to be $p_{DB}=0.75$ with $\epsilon=0.05$.

\begin{table}
\centering
\caption{Precision, Recall and F1 scores for class \textit{I} of 
EC detection.
We report results for different models trained using combinations of modalities (acoustics and lexical) with early (EF), late (LF) and decision level late fusion (DLF) using posterior probabilities.
For LF, only the best performing experiment using a fully connected neural network (FCNN) is shown.
Baseline results are included for equal priors with $p_{I}=p_{O}=0.5$ representing a fair coin toss and class priors with $p_{I}=0.066$ and $p_{O}=0.934$. 
The results are in the format: \textit{mean(std)}; computed over the five folds.}
\setlength\tabcolsep{7 pt}
\label{tbl:results}
\scalebox{0.86}{\begin{tabular}{|l|c|c|c|c|} 
\hline
 \textbf{Model} & \textbf{Features}  & \textbf{Prec-I}  & \textbf{Recall-I}  & \textbf{F1-I}  \\ 
\hline
Baseline & equal priors & 6.6  & 50.0  &  0.12 \\ 
\hline
Baseline & class priors & 6.6  & 6.6  & 0.07 \\ 
\hline
ResNet18 & MFCCs & 19.4 (5.3) & 64.6 (14.6)  & 0.29 (0.05) \\ 
\hline
ST &  WAE & ~7.6 (2.3) & 40.3 (9.0) & 0.13 (0.03) \\ 
\hline
ST & WTE & 37.9 (6.9) & 46.7 (6.4) & 0.41 (0.03) \\ 
\hline\hline
ST EF & WTE, WAE &  35.3 (6.4) & 44.3 (5.9) & 0.39 (0.04) \\
\hline\hline
FCNN LF &  logits      &   25.6 (3.6)      & 52.5 (9.2)            & 0.34 (0.01)          \\ 
\hline\hline
\textbf{DLF}  & \textbf{post. prob.} & \textbf{42.3 (5.3)}  & \textbf{51.2 (6.4)}  & \textbf{0.46 (0.05)}  \\
\hline\hline
Oracle & - & 70.6 (6.1) & 67.4 (4.6) & 0.69 (0.03) \\ 
\hline
\end{tabular}}
\end{table} 
\section{Discussion and Conclusion}\label{sc:discussion}

With a strong lexical baseline and the promising results from previous work, we were convinced that ordinary fusion strategies would help to improve our results.
The high recall on the acoustic ResNet18 system was encouraging.
However, the results for EF and LF experiments suggest that simply adding the acoustic representations, extracted from the ResNet, adds a lot of entropy that the system in its current architecture can't handle, yielding worse accuracy than the textual system.

The analysis of the extracted representations and our knowledge about the existence of acoustic cues led us to explore heuristic ways to combine the modalities.
The final DLF experiments show that the accuracy of the lexical model with its knowledge about context and content of a narrative could be improved by relying on local acoustic information in case of uncertainty.

The task of detecting EC is an important step towards a deeper understanding and better modeling of a person's emotional state; ECs can also benefit natural language understanding tasks such as dialog modeling.
Combining acoustic and lexical modalities yields higher accuracy than the uni-modal approaches to this difficult task if done the right way.
We could show that local acoustic information alone is not reliable to detect EC but helps to improve results when combined with a text-based system that captures long-range semantic relations.

The research on acoustic cues of emotion carriers is still in the initial stage.
In future work, we will look into more effective word-level acoustic representations that can be used in typical fusion approaches.

\section{Acknowledgments}
The research leading to these results has received funding from H2020 Grant Agreement 826266: COADAPT.
This work is also supported by the Bayerisches Wissenschaftsforum (BayWISS) and the German Academic Exchange Service (DAAD).

\newpage
\bibliographystyle{IEEEbib}
\bibliography{mybib}


\end{document}